\documentclass[sigconf]{acmart}
\pdfoutput=1
\usepackage{graphicx}
\usepackage{booktabs} 
\usepackage{makecell}
\usepackage{subcaption}
\usepackage{multirow}
\usepackage{array}
\usepackage{graphicx}  
\usepackage{tabularx}
\usepackage{booktabs} 
\usepackage{makecell}
\usepackage{subcaption}
\usepackage{multirow}
\usepackage{array}
\usepackage{adjustbox}
\usepackage{graphicx}
\usepackage{booktabs} 
\usepackage{makecell}
\usepackage{subcaption}
\usepackage{multirow}
\usepackage{array}
\usepackage{booktabs}
\usepackage{multirow}
\usepackage{tikz}
\usepackage{graphicx}
\usepackage{comment}
\usepackage{multirow}
\usepackage{multicol}
\usepackage{array,amsmath,fancyvrb,tabularx}
\usepackage{amssymb}
\usepackage{tikz}
\usepackage{makecell}
\usepackage{lipsum}
\usepackage{pgfplots,pgfplotstable}
\usetikzlibrary{patterns}
\pgfplotsset{compat=1.4}
\usetikzlibrary{pgfplots.groupplots}
\pgfplotsset{compat=1.11}
    \usetikzlibrary{
        pgfplots.colorbrewer,
    }
    \pgfplotsset{
        cycle list/.define={my marks}{
            every mark/.append style={solid,fill=\pgfkeysvalueof{/pgfplots/mark list fill}},mark=*\\
            every mark/.append style={solid,fill=\pgfkeysvalueof{/pgfplots/mark list fill}},mark=square*\\
            every mark/.append style={solid,fill=\pgfkeysvalueof{/pgfplots/mark list fill}},mark=triangle*\\
            every mark/.append style={solid,fill=\pgfkeysvalueof{/pgfplots/mark list fill}},mark=diamond*\\
        },
    }
\definecolor{aa}{rgb}{0.2,0.7,0.310}
\definecolor{cc}{rgb}{0.914,0.725,0.431}
\definecolor{bb}{rgb}{0.514,0.325,0.831}
\newcommand\blfootnote[1]{%
  \begingroup
  \renewcommand\thefootnote{}\footnote{#1}%
  \addtocounter{footnote}{-1}%
  \endgroup
}

\usepackage{textcomp}
\usepackage{epsfig} 
\usepackage{algorithm}
\usepackage{amsmath}
\usepackage{balance}
\usepackage{titlesec}

\renewcommand{\thefootnote}{\fnsymbol{footnote}}
\titleformat*{\subsubsection}{\large\bfseries}
\titleformat*{\section}{\LARGE\bfseries}
\titleformat*{\subsection}{\Large\bfseries}
\titleformat*{\subsubsection}{\large\bfseries}
\titleformat*{\paragraph}{\large\bfseries}
\titleformat*{\subparagraph}{\large\bfseries}
\usepackage{algpseudocode}
\algnewcommand\algorithmicforeach{\textbf{foreach}}
\algdef{S}[FOR]{ForEach}[1]{\algorithmicforeach\ #1\ \algorithmicdo}

\setcopyright{rightsretained}

\title{Creating A Neural Pedagogical Agent by Jointly Learning to Review and Assess}

\author{Youngnam Lee*}
\affiliation{Riiid! Research}
\email{yn.lee@riiid.co}

\author{Youngduck Choi*}
\affiliation{Riiid! Research, Yale University}
\email{youngduck.choi@yale.edu}

\author{Junghyun Cho}
\affiliation{Riiid! Research}
\email{jh.cho@riiid.co}

\author{Alexander R. Fabbri}
\affiliation{Yale University}
\email{alexander.fabbri@yale.edu}

\author{Hyunbin Loh}
\affiliation{Riiid! Research}
\email{hb.loh@riiid.co}

\author{Chanyou Hwang}
\affiliation{Riiid! Research}
\email{cy.hwang@riiid.co}

\author{Yongku Lee}
\affiliation{Riiid! Research}
\email{yk.lee@riiid.co}

\author{Sang-Wook Kim}
\affiliation{Hanyang university}
\email{wook@hanyang.ac.kr}

\author{Dragomir Radev}
\affiliation{Yale University}
\email{dragomir.radev@yale.edu}

\begin{document}
\begin{abstract}\label{abs}
Machine learning plays an increasing role in intelligent tutoring systems as both the amount of data available and specialization among students grow. Nowadays, these systems are frequently deployed on mobile applications. Users on such mobile education platforms are dynamic, frequently being added, accessing the application with varying levels of focus, and changing while using the service. The education material itself, on the other hand, is often static and is an exhaustible resource whose use in tasks such as problem recommendation must be optimized. The ability to update user models with respect to educational material in real-time is thus essential; however, existing approaches require time-consuming re-training of user features whenever new data is added. In this paper, we introduce a neural pedagogical agent for real-time user modeling in the task of predicting user response correctness, a central task for mobile education applications. Our model, inspired by work in natural language processing on sequence modeling and machine translation, updates user features in real-time via bidirectional recurrent neural networks with an attention mechanism over embedded question-response pairs. We experiment on the mobile education application SantaTOEIC, which has 559k users, 66M response data points as well as a set of 10k study problems each expert-annotated with topic tags and gathered since 2016. Our model outperforms existing approaches over several metrics in predicting user response correctness, notably out-performing other methods on new users without large question-response histories. Additionally, our attention mechanism and annotated tag set allow us to create an interpretable education platform, with a smart review system that addresses the aforementioned issue of varied user attention and problem exhaustion.



\end{abstract}



\maketitle

\blfootnote{*Equal contribution.}
\section{Introduction}\label{sec1}
Artificial Intelligence (AI) in education can take the form of many applications, from correctness prediction to problem suggestion and score prediction, among others. Early research in intelligent tutoring systems (ITS) focused on user/item modeling using domain-specific knowledge with the aid of education professionals. In recent years, however, fully data-driven approaches using large data have shown better results than methods exploiting domain-specific knowledge \citep{bobadilla2009collaborative, thai2010recommender}, and machine learning (ML) plays a large role in current education applications. As a result, providing personalized learning through the analysis of data has become increasingly popular. With the rise in popularity of ITS, mobile education platforms are significantly growing in size, quality, and impact on general education \cite{klassen2013requirements, west2015connected}. Mobile education applications deliver an enormous amount of education content in a flexible form while simultaneously collecting various forms of education data. 
\par
The problems we are concerned with fall under the umbrella of test preparation for mobile applications, and we focus on correctness prediction of user responses in a sequence, a task essential for downstream tasks such as predicting a user's score and suggesting the next problem a user sees. The correctness and incorrectness of user responses are denoted 1 and 0. Thus, user response data can be considered a binary ranked problem; items with the lower predicted probability are subsequently recommended first because the incorrectness of response implies some information about what the user does not know. While this approach provides a decent heuristic, one downside is that the model score does not necessarily provide an interpretable meaning of the information present in suggested problems, a problem we address later. 

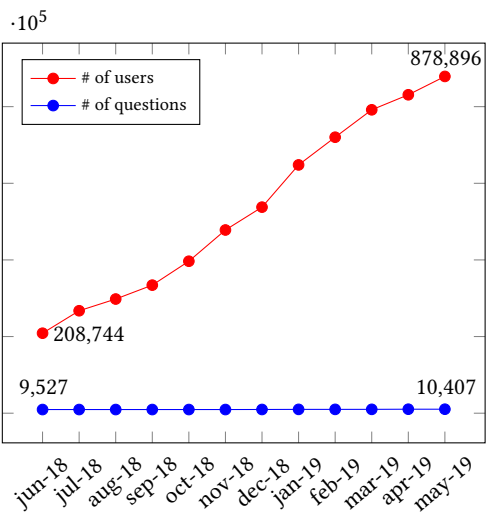
\begin{figure}[b]
\centering
\begin{tikzpicture}
 \filldraw (1.15,1.1){} node[anchor=south,fill=white,yshift=0.1cm] {208,744};
 \filldraw (5.9,4.77){} node[anchor=south,fill=white,yshift=0.1cm] {878,896};
  \filldraw (0.55,0.4){} node[anchor=south,fill=white,yshift=0.1cm] {9,527};
  \filldraw (5.9,0.4){} node[anchor=south,fill=white,yshift=0.1cm] {10,407};
\begin{axis}[
width=8cm,
xtick=data,
xticklabel style={rotate=40},
symbolic x coords={jun-18,jul-18,aug-18,sep-18,oct-18,nov-18,dec-18,jan-19,feb-19,mar-19,apr-19,may-19},
legend style={at={(0.04,0.96)},anchor=north west, font=\footnotesize, legend cell align=left}
]
\addplot[color=red, mark=*]
   coordinates {
   (jun-18,208744)(jul-18,267342)(aug-18,297928)(sep-18,334203)(oct-18,396614)(nov-18,478109)(dec-18,537761)(jan-19,647960)(feb-19,720263)(mar-19,791780)(apr-19,830994)(may-19,878896)
   };
  \addlegendentry{\# of users},

\addplot[color=blue,mark=*]
   coordinates {
   (jun-18,9527)(jul-18,9527)(aug-18,9527)(sep-18,9527)(oct-18,9527)(nov-18,9527)(dec-18,9687)(jan-19,9967)(feb-19,9967)(mar-19,10047)(apr-19,10407)(may-19,10407)
   };
  \addlegendentry{\# of questions};
\end{axis}
\end{tikzpicture}
\caption{Changes in the number of users and questions.}
\label{fig:changes_u_q}
\end{figure}


The most commonly used method for such education systems is Collaborative Filtering (CF) \cite{toscher2010collaborative, bobadilla2009collaborative}. As the basic purpose in educational platforms is to capture the relation among users and items, CF is a natural choice. However, methods such as CF are not feasible enough for real-time user modeling, as updating the user features requires re-training the model whenever new data is added. The cost of training increases exponentially as users increase. Another byproduct of the structure of educational data is the cost of suggesting problems. Under most settings, there are a finite number of resources and suggested follow-up questions which can be shown to the user. Thus, maximizing the historical sequence of problems and efficient modeling of user interactions is key to a successful educational platform, an issue we address in our method below. 
\par
In this paper, we introduce a neural pedagogical agent to model dynamic latent factors. This method is deployed on the mobile application SantaTOEIC, an English studying app in Android and iOS mobile devices with 559k users in Korea. From 2016, the service has collected more than 66M user response data as well as over 10k educational problems, as shown in Figure \ref{fig:changes_u_q}. Each problem has additionally been tagged according to topic by education experts. Regarding the characteristics of education data, we suggest a model where the item features are fixed (as shown in the figure, the number of problems has remained almost constant), but the user features are updated from responses in real-time via bidirectional recurrent neural networks \cite{lecun2015deep, wu2017recurrent}. Our model is inspired by work in NLP on learned representations such as word embeddings \cite{mikolov2013efficient, mikolov2013distributed, pennington2014glove} and machine translation \cite{cho2014properties} , and also makes sure of an attention mechanism \cite{bahdanau2014neural}  which offers an interpretable way to intelligently suggest follow-up questions, which we demonstrate with respect to the tag annotations. We perform experiments with data from SantaTOEIC, although our method can be applied in other frameworks which require a user model. Empirical results demonstrate that our neural pedagogical agent can achieve efficient real-time user modeling while also having better effectiveness in accuracy compared to existing methods. In summary, our main contributions are as follows:
\begin{itemize}
  \item We propose a scalable user model for a real-world education application.
  \item We show improved accuracy compared to existing methods tested on data from SantaTOEIC with 559k users, particularly showing improvement for new users without large question-response histories. 
  \item Our model, by embedding questions with attention-based bidirectional recurrent neural networks, allows us to create an interpretable education platform with a smart review system that addresses the lack of consistent user focus and problem exhaustion for the downstream task of problem recommendation. 
\end{itemize}
The paper is organized as follows. Section 2 reviews related work in ITS and NLP. Section 3 introduces the implementation of the service and the characteristics of the gathered data. Section 4 presents the proposed method. Section 5 demonstrates the superiority of our approach in comparison with existing approaches via experimental evaluation. Finally, Section 6 summarizes and concludes the paper.


\blfootnote{The number of users in figure 1 includes our beta test users.}



\section{Related work}\label{sec2}
AI in education (AIEd) aims to leverage AI for creating personalized educational technologies \cite{du2016artificial, conati2018ai}. Such systems aim to combine models for the target study domain, pedagogical knowledge and the student into a system which provides real-time feedback to the student. Lee et al. \cite{lee2016learning} demonstrate that fully data-driven collaborative filtering can achieve better results than classification models that exploit the domain-specific knowledge of experts. Related to this line of work, some researchers have suggested methods that adopt recommender systems for user performance prediction \cite{toscher2010collaborative, okubo2017neural}. Many recommender systems are based on the philosophy of collaborative filtering, where users with similar responses tend to respond similarly in the future. The main techniques of collaborative filtering can be divided into types: neighborhood approaches to compute the relationships between users and items  \cite{desrosiers2011comprehensive, bell2007improved} and latent factor models to transform explicit data into low-dimensional user/item latent factors by matrix factorization techniques \cite{koren2008factorization, Sal07}.
\par
With the increasing presence of data-driven machine learning applications, there has also been a call for interpretable models which both help the user gain trust in the system and improve system understanding \cite{conati2018ai, kostakos2017avoiding}, offering a level of transparency for both the user and the researcher \cite{weller2017challenges}. This call for interpretability extends not only to educational applications but more broadly to machine learning research \cite{gilpin2018explaining}. 
\par
Recent work in deep learning has benefited from advances in learned representations such as word embeddings \cite{mikolov2013efficient, mikolov2013distributed, pennington2014glove}. These embedding representations aim to capture co-occurrence relationships in the embedding space. Additionally, sequence modeling, in particular tasks such as machine translation, have shown great improvements through the adoption of neural sequence-to-sequence models \cite{cho2014properties} and subsequent introduction of attention-based models \cite{bahdanau2014neural}. Attention mechanisms helped address the problem of long-term dependencies inherent in sequence modeling and offer an interpretable model of sub-sequence importance. These mechanisms build a distribution over a given encoder representation which can be interpreted as the importance of each sub-representation at a given timestep. This has been applied to many tasks such caption generation \cite{xu2015show} and predictive modeling for healthcare \cite{choi2016retain}, and we discuss this in relation to our model below. 
Much recent work in Natural Language Processing (NLP) has aimed to address the task of multiple-choice question and answer generation \cite{kumar2016, araki2016generating} and expanded into the science domain \cite{schoenick2017}. Other work has aimed to automatically determine the quality of questions \cite{kopp2017}. While we make use of NLP techniques in modeling our task, we deal with the relationship of questions and answers to the students in the task of predicting whether or not the student will answer a given question correctly based on their history.

\section{AIEd in Action: SantaTOEIC}
\begin{figure}[t]
\centering
\includegraphics[width=0.45\textwidth]{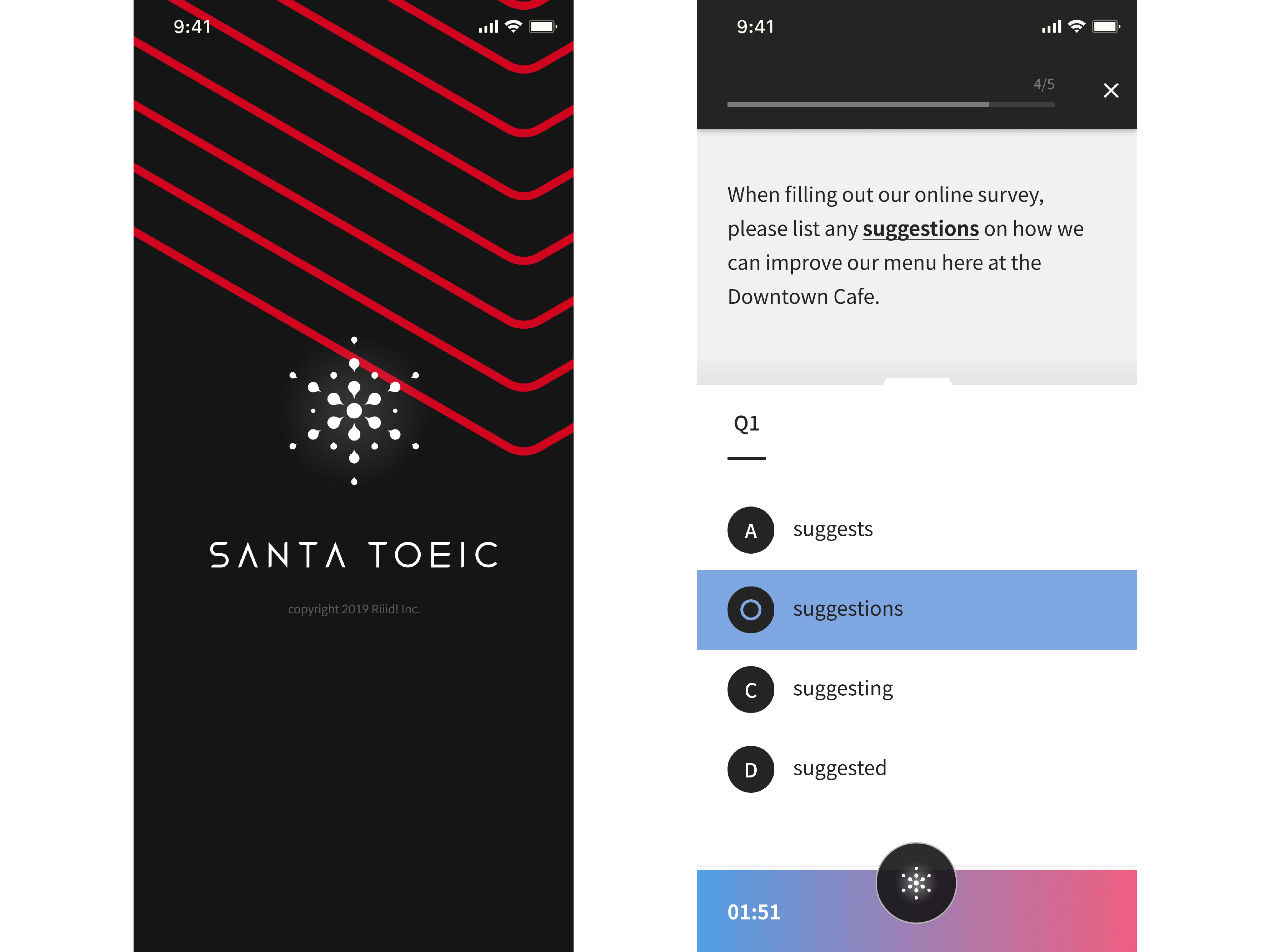}
\caption{User interface of SantaTOEIC.}
\label{ui}
\end{figure}

SantaTOEIC is an off-the-shelf AI tutor service for English education. Specifically, it aims to prepare students for the TOEIC Listening and Reading Test. This test consists of two timed sections (listening and reading) of 100 questions each, and scores range from 0 to 990, with a score gap of 5. Figure \ref{ui} shows the user interface of SantaTOEIC. Users can study English by watching lectures and reading the explanations of questions to which they responded. Currently, SantaTOEIC is available via Android and iOS applications, and over 559k users have signed up for the service.
In this section, we introduce three aspects of SantaTOEIC and education data which build an understanding of the motivations in this paper.

\subsection{User Response Correctness Prediction}
SantaTOEIC utilizes several AI techniques to optimize the learning process of users.
Newcomers to SantaTOEIC start using the platform by taking a short diagnostic test consisting of 6 problems, where each problem is chosen in real-time to maximize the accuracy of the predicted TOEIC score. As there are about 10k problems, two users typically solve different sets of problems for the initial diagnosis. 
After this diagnosis, our neural pedagogical agent then models user correct response prediction, a backbone of our platform which is used in downstream tasks such as score analytics and problem suggestion. User response correctness allows for dynamic problem suggestion in the following manner: we compute user correctness over all problems, eliminate problems with high probability and select the content based on expert heuristics. Additionally, the attention mechanism we introduce allows us to enhance our platform with a smart review system, which we discuss below. 
\par
The SantaTOEIC recommender system for problem suggestion naturally extends classical Item Response Theory (IRT). IRT is a research field for the measurement of skills based on user-item response data. Items are selected to measure based on previous results so that the response of this item would result in the most accurate diagnosis \cite{irt1, irt2}. Typically, the chosen item is the one that maximizes Fisher information. 


%


\begin{figure}[t]   
\centering
\includegraphics[width=0.45\textwidth]{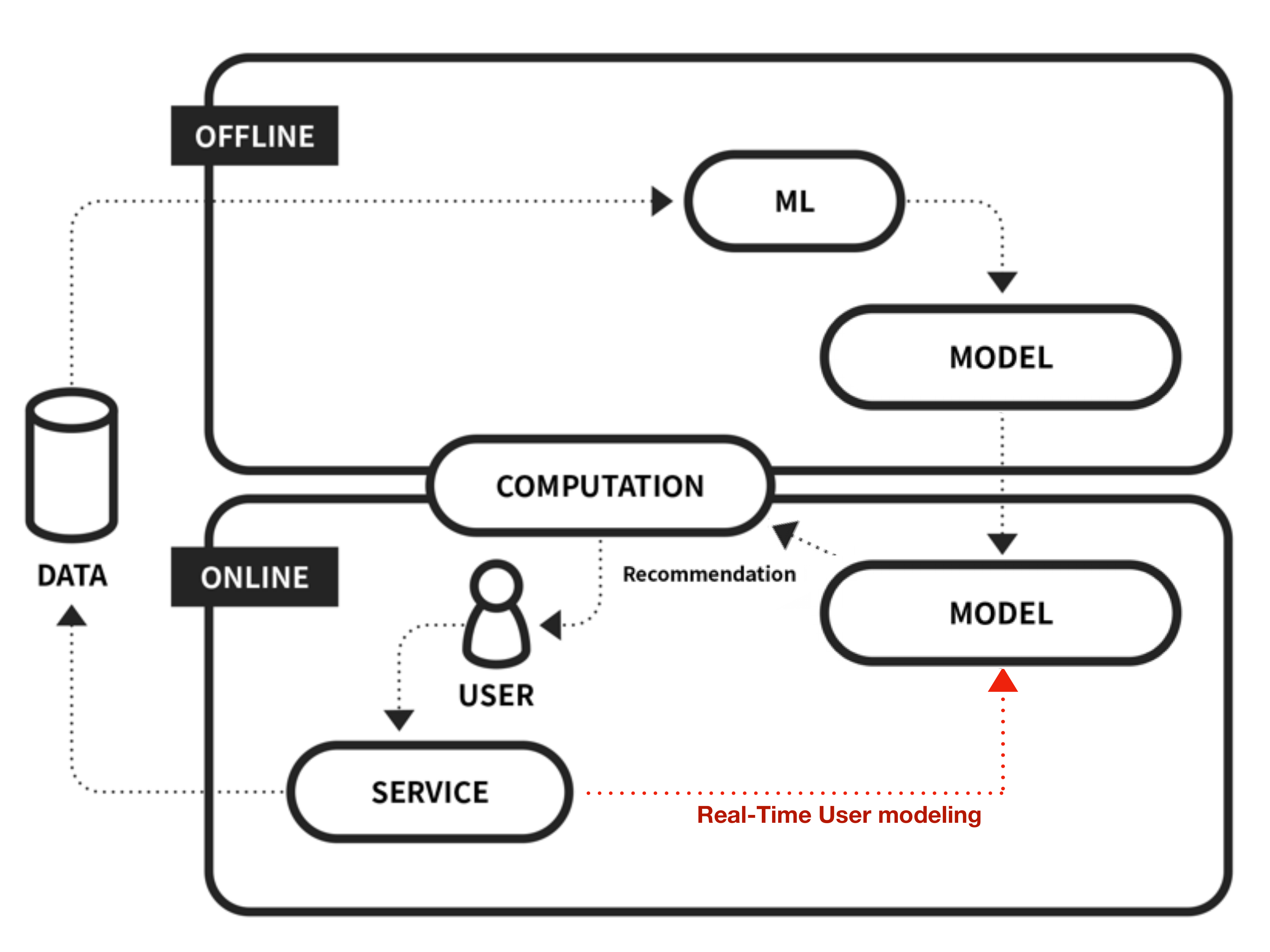}
\caption{Flow of service.}
\label{flow}
\end{figure}

\subsection{Explainable AI in Education}
Explainability, or interpretability is an important aspect of effective and trusted AI in education \cite{conati2018ai, a1}.
Explainable AI helps students better understand their comprehension levels and developmental progress over time by showing why certain educational content is recommended. Students can then learn more efficiently by understanding their weaknesses and the reason why they should learn a concept. Connected to the idea of relaying a proper explanation to the user, related research has also shown the effects of proper visual analytics and virtual agents on metacognition \cite{https://www.sciencedirect.com/science/article/pii/S1877042814009008, https://link.springer.com/article/10.1007/s11409-006-6893-0}.
Despite the fact that explainable AI can give large improvements in education efficiency, explainable AI requires reliable annotation and tags from domain experts.
The cost of this process can be large in cases when data is rare or the field is specialized.
However, the education contents in SantaTOEIC, which include about 10k problems and 500 video lectures, were tagged by a team of 20 domain experts for research/service purposes. With these high quality tags, SantaTOEIC offers explainable AI features such as the smart review systems, weakness analysis, and a virtual agent.
\subsection{Smart Review System}
\begin{figure}[t] 
\centering
\includegraphics[width=0.4\textwidth]{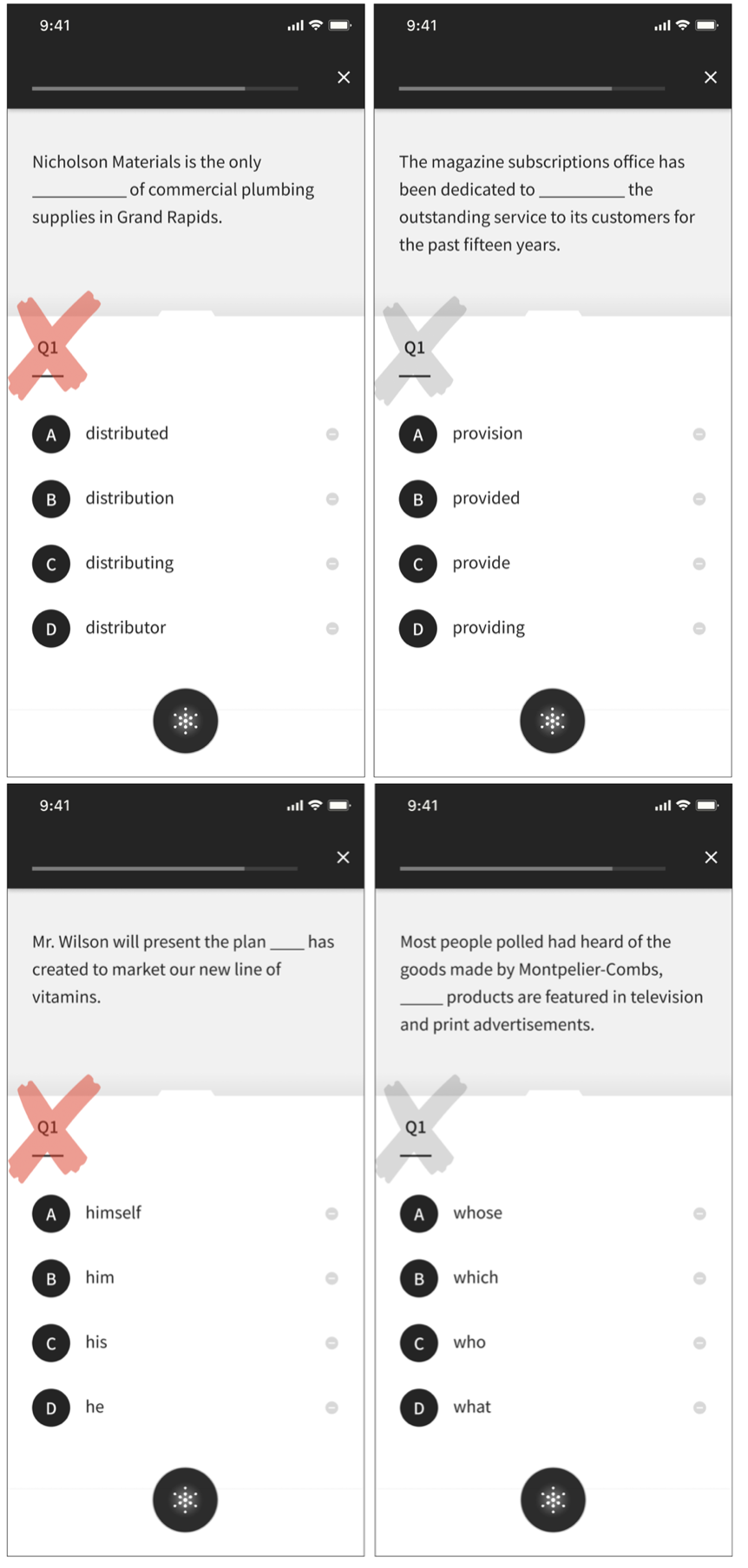}
\caption{Example of review.}
\label{att}
\end{figure}

The smart review system plays an important part in SantaTOEIC's educational platform. A notable characteristic of the education field is the high cost of education content. Educational material should be designed carefully, inspected by experts and aligned with other resources. However, due to weak concentration and motivation of students inherent in the on-the-go nature of mobile applications, no guarantees exist that a student faithfully consumes educational content in the mobile world. Using the attention mechanism mentioned above, and which we will describe fully below, helps resolve this problem. By letting students review the items with high attention important to the pedagogical agent, the smart review system of SantaTOEIC improves the total effectiveness of the system. 
Figure \ref{att} shows two cases of the smart review system that illustrate how the production system uses attention. In each case, the problem on the right is a problem whose response correctness prediction is very low according to our agent. The problem on the left side is the current most highly-attended problem that the user answered incorrectly. One can notice that the problem pairs have different vocabularies or sentence structures, but also that the problems relate to the same English concept; they both require the user to understand the correct part of speech needed to correctly answer the question. By understanding the relationship between previously-answered problems and possible suggestions, the smart review system of SantaTOEIC can maximize the user of high-cost educational content by correctly choosing the content which will maximize a user's potential.

\section{User Modeling with Bidirectional LSTM and Additive 
Attention}\label{sec4}
\begin{figure*}[t]
\centering
\includegraphics[width=0.8\textwidth]{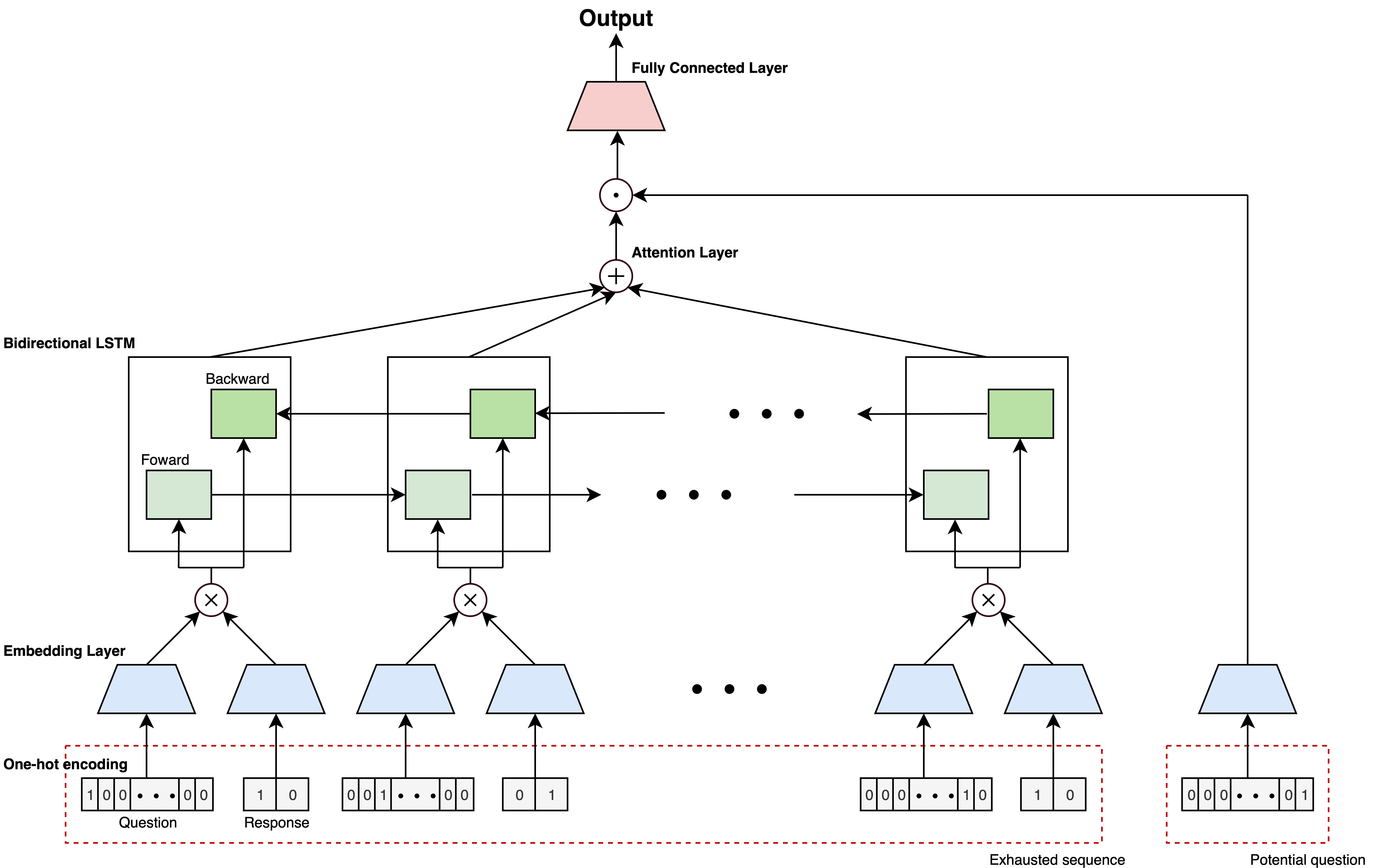}
\caption{Overview of the architecture.}
\label{architecture}
\end{figure*}

In this section, we propose a novel model based on bidirectional Long Short-Term Memory Networks (Bi-LSTM) \citep{hochreiter1997long, schuster1997bidirectional} and an additive attention mechanism \citep{bahdanau2014neural}. The overall architecture of the proposed model is outlined in Figure \ref{architecture}. First, we define the details of our problem and the representation of the data (Sec. \ref{sec41} and Sec. \ref{sec42}). Next, we introduce how our model captures user features from the user's previous responses (Sec. \ref{sec43}) and how to predict the response correctness probability from user features and question features (Sec. \ref{sec44}).

\subsection{Using Sequential Questions and Responses}\label{sec41}
Our model is designed to predict the response correctness probability given vectors $u_i$ representing user $i$ and $q_j$ representing question $j$ which may not have been answered yet. Here, we call the question to be predicted the \textit{potential question}. Given a user and potential question vector pair $u_i, q_j$, in other words, we predict the response correctness probability $p_{ij}$ as:

\begin{equation}\label{eq1}
    p_{ij} = f(u_i, q_j)
\end{equation}

We call the sequence of question-response pairs the \textit{exhausted sequence}. The user vectors $u_i$ are mapped using the exhausted questions $q_k$ and the responses $r_k \in \{0, 1\}$ to them. The details of the method will be introduced in Section \ref{sec43}. To capture the temporal properties of exhausted sequences, we add time indices to the potential questions and the exhausted sequence. Thus Eq.(\ref{eq1}) becomes:

\begin{equation}\label{eq2}
    p_{ij} = f(u_i, q_j^t)
\end{equation}
where $u_i$ at time $t$ is computed using previous responses only,
\begin{equation}\label{eq3}
    u_i = g(q^{<t}, r^{<t})
\end{equation}
where  $q^{<t}$ and $r^{<t}$ are the sequence of questions and responses exhausted before time $t$.

Once our model is trained, the model can adaptively map user vectors in the latent space without extra training. This enables our model to map new users without updating the model. 

\subsection{Data Representation}\label{sec42}
Typically, words (or other units such as characters, sentences) are embedded in vector spaces of appropriate dimensions, and these vectors are fitted to capture linguistic properties such as morphology, syntax and semantics. Similarly, we embed questions and responses respectively to some $d$-dimensional vector space from the user-item response data.

Once the model is trained, the embedded representations capture the information about the questions such as difficulty, type and other details. Work on word embedding has analyzed the resulting embedding structure, leading to the finding that embedding (King) - embedding (Man) + embedding (Woman) $\sim$ embedding (Queen) in \citep{mikolov2013linguistic}. By analyzing the expert-annotated tags, we were able to generate analogous analysis for question pairs. Table \ref{table1} shows how the tags of questions whose embeddings have high cosine similarity are closely related. Table \ref{table2} shows that subtracting and adding question vectors results in a question semantically similar to the question whose tags are found by applying the analogous set operations. 

\begin{table}[h]\label{table1}
\centering
\caption{Question synonym}
\begin{tabular}{r||l}
\toprule 
$question_{345}$& [indirect question, part2]\\
\hline
$question_{6873}$ (0.971) & [choice question,  part2]\\
$question_{5753}$(0.965) & [direct question -when,  part2]\\
$question_{362}$ (0.965) & [do/be/have question,  part2]\\
\hline
\multicolumn{2}{c}{}\\
\hline
$question_{143}$& [to infinitive, part5]\\
\hline
$question_{3875}$ (0.849) & [verb tense, part5]\\
$question_{7178}$ (0.833) & [to infinitive vs. gerunde, part5]\\
$question_{3047}$ (0.818) & [vocabulary -adverb, part5]\\
\hline
\multicolumn{2}{c}{}\\
\hline
$question_{23}$& [countable/uncountable noun, part5]\\
\hline
$question_{8}$ (0.999) & [countable/uncountable noun, part5]\\
$question_{5458}$ (0.771) & [part of speech -adjective, part5]\\
$question_{2937}$ (0.764) & [part of speech -verb, part5]\\
\bottomrule
\end{tabular}
\label{table1}
\end{table}

\begin{table}[h!]\label{table2}
\centering
\caption{Question analogy}
\begin{tabular}{lllll}
\textbf{Example 1.} & &  & & \\
\hline
$question_{11305}$&-&$question_{9420}$&+&$question_{3960}$\\
\cmidrule[1.5pt]{1-5}
double document & &  double document& & single document\\
email form    &  & email form   & &  announcement\\
announcement    &$\setminus$&  detail&$\bigcup$&detail\\
inference&  &   &&\\
implication &&    &&\\
\hline
\cline{2-5}
&&&$\approx$&$question_{10301}$\\
\cmidrule[1.5pt]{2-5}
&&single document&&single document\\
&&announcement&&announcement\\
&$=$&inference&$\approx$&inference\\
&&implication&&implication\\
&&detail &&\\

\end{tabular}
\begin{tabular}{lllll}
&&&&\\
\textbf{Example 2.} & &  & & \\
\hline
$question_{10365}$&-&$question_{4101}$&+&$question_{1570}$\\

\cmidrule[1.5pt]{1-5}
single document & &  single document& & direct question\\
email form    & $\setminus$ & email form   &$\bigcup$&  when\\
true    &&  inference&&true\\
NOT/true&  &   &&\\
\hline
&&&$\approx$&$question_{2385}$\\
\cmidrule[1.5pt]{2-5}
&&direct question&&direct question\\
&$=$&when&$\approx$&when\\
&&true&&true\\
&&NOT/true&&when vs. where\\
\cline{2-5}
\end{tabular}
\label{table2}
\end{table}

\begin{figure*}[t]
\centering
\includegraphics[width=0.9\textwidth]{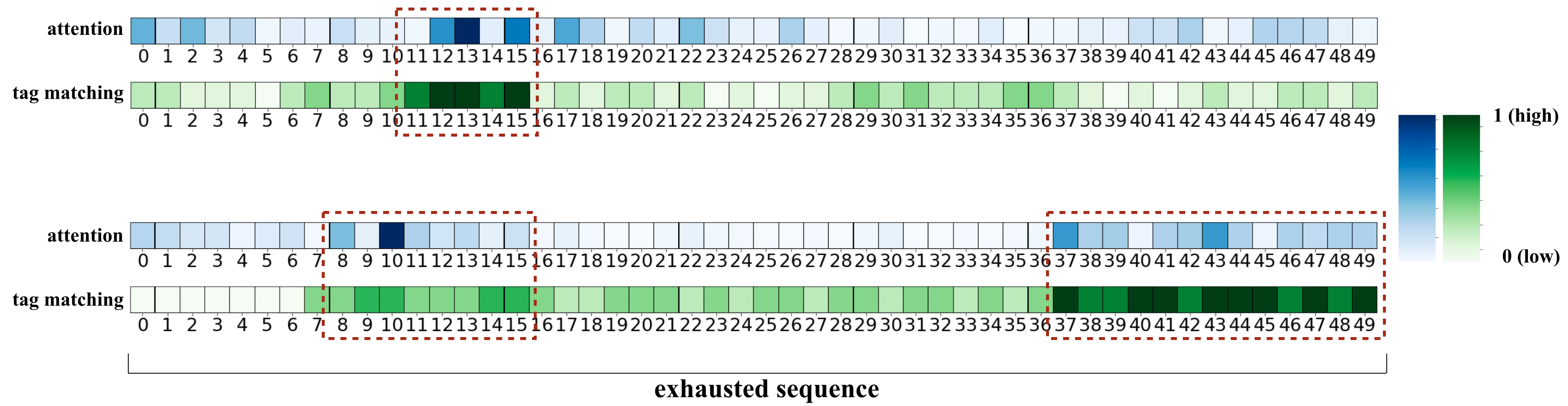}
\caption{Attention and tag matching.}
\label{fig:attention_tag}
\end{figure*}

\subsection{Modeling User Features from Exhausted Sequences}\label{sec43}
In our architecture, the user features are captured by exhausted sequences as in Eq. (\ref{eq3}). To capture user features, we use Bi-LSTM \citep{schuster1997bidirectional} and an additive attention mechanism \citep{bahdanau2014neural}. First we define the inputs for each step of the Bi-LSTM by computing the element-wise multiplication of embedded question vectors and embedded response vectors.

\begin{equation}\label{eq4}
    i^t = q^t\circ r^t
\end{equation}
where $\circ$ is the element-wise multiplication. Then, the Bi-LSTM output is defined as:

\begin{equation}\label{eq5}
    z^t = BiLSTM(i^t)
\end{equation}

\begin{figure}[h]
\centering
\includegraphics[width=0.35\textwidth]{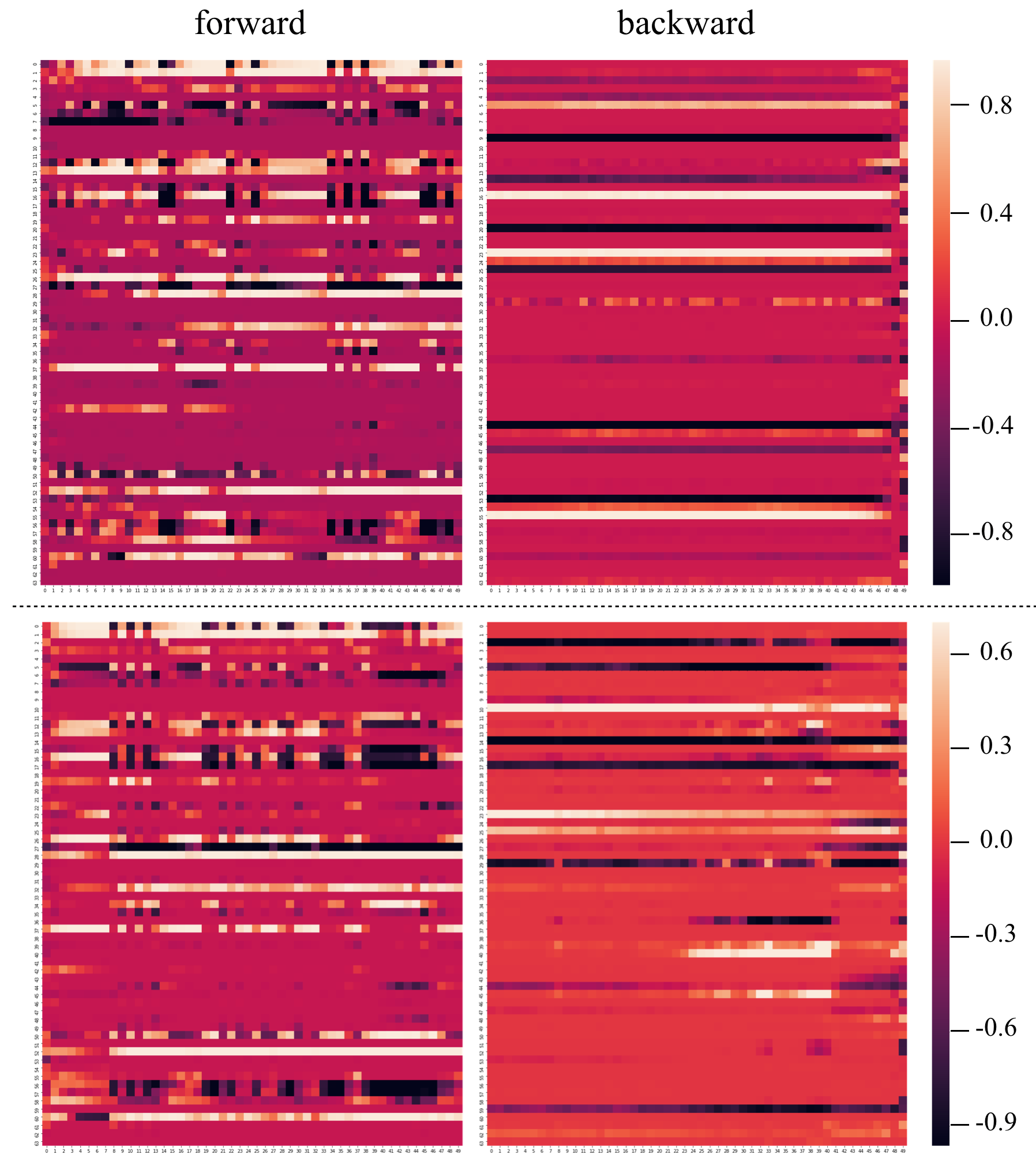}
\caption{Bidirectional LSTM output heatmap.}
\label{lstm_heatmap.}
\end{figure}

Using Bi-LSTM allows our model to track user features such as grade not only in the forward direction but also in the backward direction.
Figure \ref{lstm_heatmap.} visualizes the LSTM output for each direction. The figure shows that the output of the backward LSTM has less variation than that of the forward LSTM. This result implies that tracking users in the backward direction is more robust and suggests that understanding how the user developed in relation to earlier sessions is key to this problem.
To get better results for predicting responses of a potential questions, we construct the user vector as:

\begin{equation}\label{eq6}
    u_i = Attention({z^{<t}}, q_j^t)
\end{equation}
Then, the user vector $u_i$ contains information about the importance of question-response pairs in the exhausted sequence to predict the response correctness of potential questions.
These attention scores give an indication as to which questions that student should review in the SantaTOEIC smart review system.
In Figure \ref{fig:attention_tag}, we see that the question's attention scores are similar to the tag matching ratio calculated by dividing the intersection of exhausted questions tags and potential question tags by the number of potential question tags.
Variants of attention mechanisms such as additive attention, or dot-product attention \citep{luong2015effective, bahdanau2014neural}, can also be applied to Eq. (\ref{eq6}). Additive attention gives the highest score in our experiments. We specify Eq. (\ref{eq6}) applying additive attention, as the following:

\begin{equation}\label{eq7}
    u_i = \sum_{k=1}^{t-1}\alpha^kz^k
\end{equation}
\begin{equation}\label{eq8}
    \alpha^k = \frac{\exp(e^k)}{\sum_{j=1}^{t-1}\exp(e^j)}
\end{equation}
\begin{equation}\label{eq9}
    e^k = a(z^k, q^t)
\end{equation}
\begin{equation}\label{eq10}
    a(z^k, q^t) = v_a^T\tanh(z^kW_a + q^tU_a)
\end{equation}
where $W_a\in\mathbf{R}^{2d_l\times d_a}$, $U_a\in\mathbf{R}^{d \times d_a}$, $v_a\in\mathbf{R}^{d_a}$ are the weight matrices. $d_l$ and $d_a$ are the dimension of LSTM and attention embedding vectors.

\subsection{Predicting Correctness}\label{sec44}

The response correctness probability of potential questions are predicted using the embedded vector of potential questions and user features. The details of the method are introduced in the previous section. We concatenate user and potential question vectors to predict correctness probabilities and feed the vectors to four fully-connected layers. 
Thus the final output is defined as:

\begin{equation}\label{eq11}
    p_{ij} = f(u_i, q_j^t)
\end{equation}
where $f$ is the network composed by fully connected layers and the sigmoid function.

\pagebreak
\begin{table}[b]
\small
\centering
\caption{Model hyper-parameters}
\begin{tabular}{c||c}
\toprule 
Type & Tuning details\\
\hline
Question embedding size & 128\\
Response embedding size & 128\\
Bi-LSTM size & 128 $\rightarrow$ 128 $\rightarrow$ 128\\
Attention (source, target) & (exhausted sequence, potential question)\\
Attention type & additive \\
Attention layer size & 256 \\
FC layer size & 512 $\rightarrow$ 256 $\rightarrow$ 128\\
\bottomrule
\end{tabular}
\label{table:hyperparameter}
\end{table}

\begin{table*}[tb]
\centering
\caption{The F1, AUC, and ACC for the whole sequence (above) and the sub-sequence from response 1 to 50 (below)}
\begin{tabular}{c||cccccc|c}
\toprule 
Metric&MF&MLP&NeuMF&RRN&AutoRec&DeepRec&Ours\\
\hline
F1-score&0.7673&0.7813&0.7764&0.7898&0.7708 & 0.7989&\textbf{0.8114}\\
AUC&0.7163&0.7363&0.7380&0.7323&0.7137  & 0.7282&\textbf{0.7661}\\
ACC&0.6822&0.7033&0.7011&0.7093&0.6913  & 0.7093&\textbf{0.7304}\\
\bottomrule
\multicolumn{8}{c}{}\\
\toprule 
Metric&MF&MLP&NeuMF&RRN&AutoRec&DeepRec&Ours\\
\hline
F1-score&0.5865&0.6612&0.6547&0.6673&0.4296 & 0.5869&\textbf{0.6956}\\
AUC&0.6114&0.7268&0.7288&0.7404&0.6417  & 0.7071&\textbf{0.7573}\\
ACC&0.5627&0.6628&0.6629&0.6742&0.5839  & 0.6402&\textbf{0.6887}\\
\bottomrule
\end{tabular}
\label{table:result}
\end{table*}

\section{Experiments}
Here we introduce our dataset, training settings as well as experimental results and analysis. 
\subsection{Dataset}
We use the SantaTOEIC dataset, which is a set of user responses of multiple-choice questions collected over the last four years in the SantaTOEIC service (Android and iOS). The main features of the user-question response data are the following four columns: user id, question id, correct/incorrect, and timestamp. Our label is the response $\in \{0, 1\}$ correctness of a user responding to a particular question. In the dataset, there are 559,695 users who solved more than one problem. The size of the question set is 13,774. The total row count of the dataset is 66,182,925. The dataset is split into three parts: train set (455,001 users, 53,622,892 responses), validation set (50,556 users, 5,852,702 responses), and test set (54,138 users, 6,705,331 responses).

\subsection{Training Details}
Our model hyper-parameters are shown in Table \ref{table:hyperparameter}. We train the model from scratch, where the weights are initialized using Xavier uniform initialization \citep{xavier_unifom}. We use the Adam optimizer \cite{kingma2014adam} with default parameters (learning rate=0.001, beta1=0.9, beta2=0.999, and epsilon=1e-08). We save the model parameters with the best results on the validation set and then evaluate the performance of the model using the test set.

\subsection{Experimental Results}
We evaluated our model by comparing with the current state-of-the-art (SOTA) approaches: matrix factorization (MF) \citep{lee2016learning, koren2008factorization}, multilayer perceptron (MLP) \citep{zhang2019deep}, NeuMF \citep{he2017neural}, recurrent recommender networks (RRN) \citep{wu2017recurrent}, AutoRec \citep{sedhain2015autorec}, and DeepRec \citep{kuchaiev2017deeprec}. In our experiments, we selected three performance metrics: F1-score, the area under the curve (AUC) of the receiver operating characteristic, and accuracy (ACC). F1-score is the harmonic mean of precision and recall. AUC shows the sensitivity (recall) against 1-specificity. Sensitivity (resp. specificity) indicates the ratio of the number of correct predictions to the total number of responses with an incorrect label (resp. correct label). ACC means the ratio of the number of correct predictions to the total responses.


Table \ref{table:result} presents the overall results of our evaluation. It shows that our model outperforms the others.
Also, the performance of our model is the best even when a user consumes fewer questions, which is shown in Table \ref{table:result} (below). This is one of the most important factors in the user experience aspect, providing users with personalized services as quickly as possible.

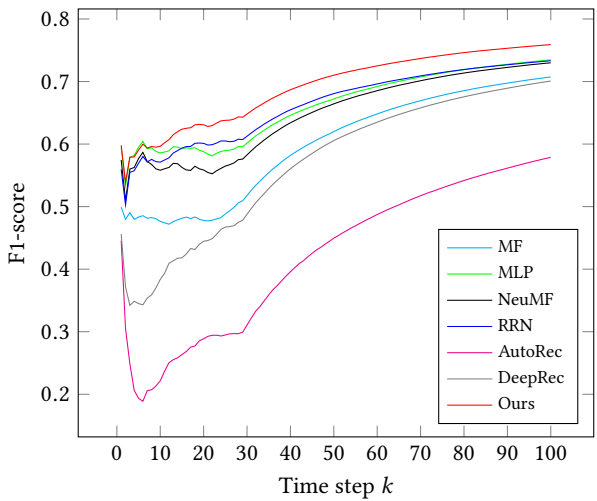
\begin{figure}[tb]
\centering
\begin{tikzpicture}
\begin{axis}[
legend pos=south east,
ylabel={F1-score},
xlabel={Time step \textit{k}},
ytick={0.2, 0.3, 0.4, 0.5, 0.6, 0.7, 0.8},
xtick={0, 10, 20, 30, 40, 50, 60, 70, 80, 90, 100},
legend style={legend cell align=left, font=\fontsize{7}{10}\selectfont}
]
\addplot[color=cyan]
   coordinates {
(1, 0.49921210211156636)(2, 0.47970072585503565)(3, 0.4903656741079179)(4, 0.4795906727254531)(5, 0.483304108509493)(6, 0.4855135165781164)(7, 0.48144825295676663)(8, 0.48246745088927095)(9, 0.48067190312290353)(10, 0.4764566055986762)(11, 0.47404771122823314)(12, 0.4718858758538813)(13, 0.47606558578975755)(14, 0.47889793574872386)(15, 0.48139723988824085)(16, 0.4832063735944277)(17, 0.4808377038042728)(18, 0.48339821540768113)(19, 0.4798152022933376)(20, 0.4778565045703518)(21, 0.4771999515565706)(22, 0.4783198566693834)(23, 0.48052349618553086)(24, 0.4824083371428372)(25, 0.4884069435508895)(26, 0.49333292666831885)(27, 0.49890295913322646)(28, 0.5060968829699105)(29, 0.5093111227861941)(30, 0.5178470462946353)(31, 0.5261570085587041)(32, 0.5340830053648465)(33, 0.5407881576424497)(34, 0.547735359758543)(35, 0.5542763211693168)(36, 0.5605644201191203)(37, 0.5659761803895562)(38, 0.5717320945962889)(39, 0.5770110184301736)(40, 0.5819189699593698)(41, 0.5863809155291495)(42, 0.5908118964302101)(43, 0.5948091452353923)(44, 0.5988592638610517)(45, 0.6026026136675693)(46, 0.6063647090591996)(47, 0.6099304946802345)(48, 0.6133503171624864)(49, 0.6163795802276159)(50, 0.6199530402483024)(51, 0.6233186071278682)(52, 0.6266178906777304)(53, 0.6295719976170457)(54, 0.6324705506249201)(55, 0.6352859024272127)(56, 0.6380773494089204)(57, 0.6405819616330934)(58, 0.6431470023343371)(59, 0.6456822939840561)(60, 0.6482123270298145)(61, 0.6506340546453507)(62, 0.6530881706866276)(63, 0.6553503808740053)(64, 0.6575108131341997)(65, 0.6596456352228579)(66, 0.6617372449901862)(67, 0.6636748886549494)(68, 0.665630350559118)(69, 0.6675300102309225)(70, 0.6694096789603697)(71, 0.6712436309701084)(72, 0.6730217981494202)(73, 0.6746359388404485)(74, 0.6763726413664928)(75, 0.6779811631057132)(76, 0.6795902030172085)(77, 0.681161082244363)(78, 0.6825267235547263)(79, 0.6840098262134023)(80, 0.6853797189924458)(81, 0.686719139024087)(82, 0.6880758591228587)(83, 0.6894158614986737)(84, 0.690650072602434)(85, 0.691839669009976)(86, 0.693031520064332)(87, 0.6941231772546201)(88, 0.6952888037390311)(89, 0.6963803214514837)(90, 0.697470465462779)(91, 0.6985765442906657)(92, 0.6995877830715213)(93, 0.7006559145231173)(94, 0.7016694179849924)(95, 0.7027159446843256)(96, 0.7036738026657235)(97, 0.7045631534465137)(98, 0.7055447039781121)(99, 0.7064433046191355)(100, 0.7073174680641722)
   };
  \addlegendentry{MF};
\addplot[color=green]
   coordinates {
(1, 0.5948504508915821)(2, 0.5329842451045769)(3, 0.5784391981417737)(4, 0.581821710694014)(5, 0.5949560143337423)(6, 0.6041633897819015)(7, 0.5924542061384303)(8, 0.5942015178735464)(9, 0.5882242737255702)(10, 0.5858773320144378)(11, 0.5872909752997173)(12, 0.5892163853453616)(13, 0.5958196742881717)(14, 0.5951653045074264)(15, 0.5923655005215708)(16, 0.5941072908089657)(17, 0.5931034467969593)(18, 0.5942239137774543)(19, 0.5902435847771296)(20, 0.5879708504747951)(21, 0.583600916223099)(22, 0.58123307900925)(23, 0.5853525096314319)(24, 0.5882935878428097)(25, 0.5897076666262319)(26, 0.5902909618088826)(27, 0.5920865615275012)(28, 0.5960710177208534)(29, 0.595739628199864)(30, 0.6016611808737335)(31, 0.6074680396979811)(32, 0.613032671976156)(33, 0.6176545444969368)(34, 0.6224636597470746)(35, 0.6268838207452772)(36, 0.6313691321019583)(37, 0.6350048192863083)(38, 0.6390447057772205)(39, 0.6426929887088364)(40, 0.6460616273528381)(41, 0.6490493907940774)(42, 0.652096611114189)(43, 0.6548791382437084)(44, 0.6576352844311274)(45, 0.6602202690542374)(46, 0.6628855984661166)(47, 0.6652767065530166)(48, 0.6676599991184989)(49, 0.6696437516871977)(50, 0.6722370892879103)(51, 0.6745761805594686)(52, 0.6769653060038441)(53, 0.6790559223006618)(54, 0.6811093776392022)(55, 0.683073481618444)(56, 0.6850263777351483)(57, 0.6867180503913928)(58, 0.6885149460194783)(59, 0.6902929256916588)(60, 0.692106897530368)(61, 0.6938221091121781)(62, 0.6955342209359974)(63, 0.6971954356622292)(64, 0.6987401004752456)(65, 0.700279045855932)(66, 0.7017430139609305)(67, 0.7031147381284006)(68, 0.7045858228550886)(69, 0.7059320123538525)(70, 0.707306179186368)(71, 0.7086621009036133)(72, 0.7099480174911122)(73, 0.7111173049390196)(74, 0.7124227525339384)(75, 0.7135672241999799)(76, 0.7147453878046853)(77, 0.7158903223678911)(78, 0.7169043130320237)(79, 0.7179937520300526)(80, 0.7189534883418638)(81, 0.719928205714169)(82, 0.720842841439251)(83, 0.7218125885724663)(84, 0.722682633133966)(85, 0.7235187676425433)(86, 0.7243882507128672)(87, 0.7251419582963144)(88, 0.7259724689081284)(89, 0.7267570597039646)(90, 0.7275341846837294)(91, 0.7283165626864322)(92, 0.7290647786225657)(93, 0.7298408523450144)(94, 0.7305733756931588)(95, 0.7313109602026223)(96, 0.7320381622165675)(97, 0.7326453556615112)(98, 0.733369756854909)(99, 0.734002778663585)(100, 0.7346394384204425)
   };
  \addlegendentry{MLP},

\addplot[color=black]
   coordinates {
(1, 0.5746675758608933)(2, 0.5098335622103998)(3, 0.5600235398278893)(4, 0.5625316334822845)(5, 0.576779975169601)(6, 0.5866847767286115)(7, 0.5721010676614169)(8, 0.5675018468745943)(9, 0.5603578675538462)(10, 0.5581898317908133)(11, 0.5604924571518101)(12, 0.5626286538431172)(13, 0.5691273288615277)(14, 0.5686074790235954)(15, 0.5626628110918143)(16, 0.5586498354500692)(17, 0.5579008908054196)(18, 0.5645372713370933)(19, 0.5604715428841228)(20, 0.5585511088586228)(21, 0.5544720126425876)(22, 0.5524821885677377)(23, 0.5574097951176755)(24, 0.5611768340553361)(25, 0.5647513351709744)(26, 0.5670173141744276)(27, 0.569865432636294)(28, 0.5751332770711303)(29, 0.5759782015226569)(30, 0.5826959386841896)(31, 0.5893016255231565)(32, 0.5956842030972138)(33, 0.6011701448735695)(34, 0.606753127678149)(35, 0.6118909754075536)(36, 0.6169719659501133)(37, 0.6213014480134532)(38, 0.6259422383175168)(39, 0.6301452734175259)(40, 0.6340447603784518)(41, 0.6375219546142775)(42, 0.6410833600826062)(43, 0.6443288262071991)(44, 0.6475219626099317)(45, 0.6505815709910561)(46, 0.6536467770958849)(47, 0.6565544780788358)(48, 0.6592480124175043)(49, 0.6616148948600725)(50, 0.6643168934678768)(51, 0.6667967031350005)(52, 0.6692775559542805)(53, 0.6714793730188442)(54, 0.6736769548935593)(55, 0.6757887808361622)(56, 0.6778663691542129)(57, 0.6796620687275671)(58, 0.6815701947488545)(59, 0.6834772808787557)(60, 0.685387455803711)(61, 0.6871658109701092)(62, 0.6889417539541325)(63, 0.6907050189692086)(64, 0.6923219540163377)(65, 0.6939439063674919)(66, 0.6954254367125563)(67, 0.6968665649764657)(68, 0.698405804520572)(69, 0.6998139770910926)(70, 0.7012811682913787)(71, 0.7027029950083183)(72, 0.704083245681158)(73, 0.7052850791354704)(74, 0.7066481029768377)(75, 0.7078778465746487)(76, 0.7091013850575743)(77, 0.7103128742228013)(78, 0.71137553318409)(79, 0.712531354186714)(80, 0.713587956178986)(81, 0.7146162824079793)(82, 0.7156035495836125)(83, 0.716581443871688)(84, 0.7175048336315474)(85, 0.7183752727340682)(86, 0.7192857828999113)(87, 0.7200895321536254)(88, 0.7209399585575946)(89, 0.7217771149813275)(90, 0.7225918861996885)(91, 0.7234180077935168)(92, 0.7241917406870013)(93, 0.7249948843183478)(94, 0.7257769933870305)(95, 0.7265503512918507)(96, 0.7272763399608991)(97, 0.7279105684036312)(98, 0.7286721166861908)(99, 0.7293277639881393)(100, 0.7299876512395335)
   };
  \addlegendentry{NeuMF};
 
  \addplot[color=blue]
   coordinates {
(1, 0.5596219364765359)(2, 0.5027865648809395)(3, 0.5548085266725012)(4, 0.5574636892223906)(5, 0.5694117377452559)(6, 0.580120412271631)(7, 0.5714162055956523)(8, 0.5755725877480433)(9, 0.5719339975483635)(10, 0.5710895810190918)(11, 0.5745170982474835)(12, 0.5779877621534245)(13, 0.5856508037099929)(14, 0.589623676876235)(15, 0.5935414498400978)(16, 0.5959743270637324)(17, 0.5962181558215986)(18, 0.601209031407396)(19, 0.601668472290183)(20, 0.6011390248355039)(21, 0.5983942122734834)(22, 0.5990655074277468)(23, 0.6017136252226127)(24, 0.6050001356902038)(25, 0.6049427313380786)(26, 0.6037718805749267)(27, 0.6044858413310296)(28, 0.6078201457646738)(29, 0.6072115151884137)(30, 0.6125234763143477)(31, 0.6177792844150335)(32, 0.6230068156724494)(33, 0.6274898186233433)(34, 0.6321378689247173)(35, 0.6363579469581186)(36, 0.6405041371209417)(37, 0.6441119239558507)(38, 0.648078073995798)(39, 0.6515900447261913)(40, 0.6548119720149006)(41, 0.6575954404514255)(42, 0.6604705612850562)(43, 0.6630699623198547)(44, 0.6658966841053502)(45, 0.6685967346616921)(46, 0.6713050412321824)(47, 0.6739479279439515)(48, 0.6763422474605764)(49, 0.6785079498398077)(50, 0.6806679065780521)(51, 0.6825700853683987)(52, 0.6843947583684087)(53, 0.6858229335140206)(54, 0.687353912275058)(55, 0.68881034179231)(56, 0.6904624680137197)(57, 0.691801876376147)(58, 0.6932284170845762)(59, 0.694643708757806)(60, 0.6961745223145563)(61, 0.6976476389554843)(62, 0.6991527975952654)(63, 0.7005237795567831)(64, 0.7017731099571706)(65, 0.7029888194013174)(66, 0.7042053887121414)(67, 0.7053871474269376)(68, 0.7066064842516282)(69, 0.7077658177592555)(70, 0.7089685784245159)(71, 0.7101493702326513)(72, 0.7112937147049799)(73, 0.712273703248395)(74, 0.7134081878975822)(75, 0.7144310833113681)(76, 0.7155050836366503)(77, 0.7164997513229754)(78, 0.7173954331872966)(79, 0.7183622545275352)(80, 0.7192610072118517)(81, 0.7201136589627231)(82, 0.7209496177214706)(83, 0.7217601390636602)(84, 0.7225037594358303)(85, 0.7232336892161753)(86, 0.723984628175587)(87, 0.7246174641885187)(88, 0.7253708722997058)(89, 0.7260781513028401)(90, 0.7267712822323871)(91, 0.7274589012065311)(92, 0.7281344751931702)(93, 0.728851870689708)(94, 0.7295386687209294)(95, 0.7301827431232555)(96, 0.7308300366256822)(97, 0.7313503679005083)(98, 0.7320059672695687)(99, 0.7326039172376206)(100, 0.7332186397013517)
   };
  \addlegendentry{RRN};
  \addplot[color=magenta]
   coordinates {
(1, 0.44551605363513347)(2, 0.30503931589109556)(3, 0.24883497937939392)(4, 0.20614046462632257)(5, 0.19348806593814377)(6, 0.18881737453608202)(7, 0.20572971981224092)(8, 0.20716528641831145)(9, 0.21307628127501876)(10, 0.2214202379527431)(11, 0.23674437340051294)(12, 0.25027237132542984)(13, 0.2552146156214301)(14, 0.25830393569989457)(15, 0.26344467865120397)(16, 0.2683338508462479)(17, 0.2753175784852008)(18, 0.2771697944302234)(19, 0.2853377390762943)(20, 0.288742489612574)(21, 0.29261527869443305)(22, 0.294249083013789)(23, 0.2943607818205665)(24, 0.2931431723291697)(25, 0.2948014360032766)(26, 0.2965539907954657)(27, 0.2969795753285393)(28, 0.2968346948403825)(29, 0.2992845054956629)(30, 0.3106678862901322)(31, 0.32194036567210604)(32, 0.3327874395874754)(33, 0.3404673134760288)(34, 0.3497495577352709)(35, 0.3588265953139594)(36, 0.3673652861239987)(37, 0.3739829236723713)(38, 0.38158663349723326)(39, 0.3890228777508879)(40, 0.3959430659676777)(41, 0.40246147387571507)(42, 0.40871492211630517)(43, 0.4134851708214691)(44, 0.4196737791291926)(45, 0.4253016978888062)(46, 0.43071423087433425)(47, 0.43496130636832153)(48, 0.4399279612063616)(49, 0.4446433489043528)(50, 0.44955939912371795)(51, 0.45353210397074917)(52, 0.4578846696848831)(53, 0.46170815741105975)(54, 0.4657803681307795)(55, 0.46932310068762745)(56, 0.4732160496802557)(57, 0.4768290407649289)(58, 0.4804726839938484)(59, 0.48378600692460383)(60, 0.48743538583874596)(61, 0.49054109722388217)(62, 0.49396340179245685)(63, 0.4969099005891096)(64, 0.49990906612645075)(65, 0.5027793382137502)(66, 0.5059128252171203)(67, 0.5088538474259141)(68, 0.5119354659754564)(69, 0.5146403161752728)(70, 0.517453237701676)(71, 0.5200583029260566)(72, 0.5227438145960727)(73, 0.5249888574707079)(74, 0.5275898519454664)(75, 0.5300265214608123)(76, 0.5324992391097576)(77, 0.5347730880695793)(78, 0.5371235160228783)(79, 0.5394407915734599)(80, 0.5417545759203647)(81, 0.5438972243726711)(82, 0.5460199151996934)(83, 0.5479943827220527)(84, 0.5500888773727617)(85, 0.5521384039730701)(86, 0.5541202224851377)(87, 0.5559431203760637)(88, 0.5577914511742819)(89, 0.5595455672884625)(90, 0.5615158526583333)(91, 0.5633298504276546)(92, 0.5652566632247036)(93, 0.5670161405836615)(94, 0.568855360017768)(95, 0.5706189557562482)(96, 0.5723580139095797)(97, 0.5739433487026518)(98, 0.5756313842195768)(99, 0.577182631931888)(100, 0.5788050493830926)
   };
  \addlegendentry{AutoRec};
  \addplot[color=gray]
   coordinates {
(1, 0.4561664190193165)(2, 0.3715617855989789)(3, 0.3421507998736463)(4, 0.348578162242144)(5, 0.3449113193538652)(6, 0.343007454011981)(7, 0.3535785170963161)(8, 0.35891042960717773)(9, 0.3700284347039965)(10, 0.38356959204327246)(11, 0.3933441904908252)(12, 0.4091860530232984)(13, 0.4133705892563943)(14, 0.4174966673825336)(15, 0.4181317910820051)(16, 0.4241446998904979)(17, 0.4329625015483897)(18, 0.4314309055850031)(19, 0.4396320467445167)(20, 0.44466255266939914)(21, 0.4458209477096765)(22, 0.4489412731489057)(23, 0.4558366375470984)(24, 0.46378892476687655)(25, 0.467264380431737)(26, 0.4679985760905749)(27, 0.46947520687100674)(28, 0.47501831709481124)(29, 0.47818153813094155)(30, 0.48774354927491265)(31, 0.4971803813534467)(32, 0.5060498776018151)(33, 0.513788406153183)(34, 0.5212844727857221)(35, 0.528759943261506)(36, 0.5358678201168766)(37, 0.5422330678238786)(38, 0.5487390142792004)(39, 0.5548128100474744)(40, 0.5605639117722802)(41, 0.5658412650889955)(42, 0.5709798960479174)(43, 0.5757356874701302)(44, 0.5805348197977556)(45, 0.5850373138550903)(46, 0.5895349293582557)(47, 0.5938901644985785)(48, 0.5979625928857537)(49, 0.6017568994906474)(50, 0.6055481783848752)(51, 0.6089143300520088)(52, 0.6122017340705829)(53, 0.6151851027789808)(54, 0.6182991325874568)(55, 0.6212688377456934)(56, 0.6242913311013506)(57, 0.626963963873136)(58, 0.6296841989105156)(59, 0.6322913106938518)(60, 0.6349776312904022)(61, 0.637591991679977)(62, 0.6401897329865288)(63, 0.6426664818323801)(64, 0.6449343434615494)(65, 0.6472462323444336)(66, 0.6494912793257117)(67, 0.6516430127248288)(68, 0.6537919138265655)(69, 0.6558128079490246)(70, 0.6578754097179511)(71, 0.659871204916703)(72, 0.66185314882352)(73, 0.6636590350610388)(74, 0.665499547960643)(75, 0.6672232574742166)(76, 0.6690062394161016)(77, 0.6707482410141054)(78, 0.6723218254413865)(79, 0.674004617214302)(80, 0.6755540032754311)(81, 0.6770970766359717)(82, 0.6785713500683894)(83, 0.6800280490471592)(84, 0.6814742795669335)(85, 0.6828350017199667)(86, 0.6842172390455374)(87, 0.6854789480737974)(88, 0.6867595058840482)(89, 0.6880347448619586)(90, 0.6893125531864635)(91, 0.6905765058433482)(92, 0.6917899147115384)(93, 0.6930336190050611)(94, 0.6942295707966007)(95, 0.6953998983777304)(96, 0.6965088958806311)(97, 0.6975782062255437)(98, 0.6987067614159098)(99, 0.6997467604588468)(100, 0.7007984786302636)
   };
  \addlegendentry{DeepRec};
    \addplot[color=red]
   coordinates {
(1, 0.5981663500454283)(2, 0.5392238791151441)(3, 0.579263344237091)(4, 0.579331842481901)(5, 0.5903989882324678)(6, 0.5998006053079278)(7, 0.5936848249776486)(8, 0.5960951787789781)(9, 0.5950754545583263)(10, 0.5963630866212207)(11, 0.6021523099674592)(12, 0.6070216678211756)(13, 0.6150093395540162)(14, 0.619845154801994)(15, 0.6237768482167669)(16, 0.6252414968944748)(17, 0.6261182620912468)(18, 0.630714621081458)(19, 0.6315558784359943)(20, 0.6307547940313167)(21, 0.6280964819437583)(22, 0.6297020721594367)(23, 0.6336468012198132)(24, 0.6371042636167924)(25, 0.6379965892636332)(26, 0.6382418176714809)(27, 0.6399108328560922)(28, 0.6434231257382691)(29, 0.6429622075119543)(30, 0.6480196771584689)(31, 0.6529800931734717)(32, 0.6577252891168368)(33, 0.6618216909708547)(34, 0.6660376659435008)(35, 0.670018082682466)(36, 0.6738101198326493)(37, 0.6771635430215198)(38, 0.6807013672601566)(39, 0.6838745310837555)(40, 0.68683826460878)(41, 0.6894919384430785)(42, 0.6921685603681432)(43, 0.6946707883982248)(44, 0.6971957343912254)(45, 0.6995464607989841)(46, 0.7019077622445739)(47, 0.7041800596304385)(48, 0.706300482085227)(49, 0.7081323277233992)(50, 0.7101178118447184)(51, 0.711868222456709)(52, 0.7136116990199337)(53, 0.7151153644684796)(54, 0.7166751496778571)(55, 0.7181026654022329)(56, 0.7196149368119438)(57, 0.7208922283412749)(58, 0.7222290214648425)(59, 0.7236378179511419)(60, 0.7250021784210533)(61, 0.726342903809982)(62, 0.7276641088493048)(63, 0.7289105032103927)(64, 0.7300987682524752)(65, 0.7312751032067416)(66, 0.7324174824858865)(67, 0.7334734990881457)(68, 0.7346903811639798)(69, 0.7357299033885347)(70, 0.7368481101453788)(71, 0.7379331887409885)(72, 0.7389603230641297)(73, 0.7398254195446685)(74, 0.7408504973650687)(75, 0.7417817535587949)(76, 0.7427215883022922)(77, 0.7436880872635414)(78, 0.7444965100324058)(79, 0.7453935345355795)(80, 0.7462168311691189)(81, 0.7469696795469112)(82, 0.7477540928450251)(83, 0.7485257032060367)(84, 0.7492414829222842)(85, 0.7498686635247371)(86, 0.7505704046766138)(87, 0.7512226310591743)(88, 0.7518721560477676)(89, 0.7525213667846785)(90, 0.7531467487256539)(91, 0.7538008891358041)(92, 0.7544154664953965)(93, 0.7550399209292246)(94, 0.755663739413541)(95, 0.7562781868857759)(96, 0.7568631795136295)(97, 0.7573421629164475)(98, 0.7579495988498284)(99, 0.7584653582024122)(100, 0.7590234887403404)
   };
  \addlegendentry{Ours};
\end{axis}
\end{tikzpicture}
\caption{The F1-score by timestep.}
\label{fig:timestep}
\end{figure}

To illustrate how model performances improve on each response step, we evaluate models at each step. For example, the F1-score at step 2 is the F1-score at the moment a user responded to the second given question. Figure \ref{fig:timestep} shows the result. Our model outperforms other models on most time steps and evaluations metrics such as AUC and ACC. Particularly, our model shows an F1-score of 0.7 after the 44th question is answered, while other methods show similar performance after at least the 64th question. One interesting result is that the F1-score decreases before the 10th question. This may be explained by the fact that the SantaTOEIC service provides the first 10 questions to maximize cross-entropy for the initial assessment. Afterwards, SantaTOEIC recommends the most effective question to the user in the education aspect. 

\subsection{Model Analysis}
\begin{figure*}[t]
\centering
\includegraphics[width=0.75\textwidth]{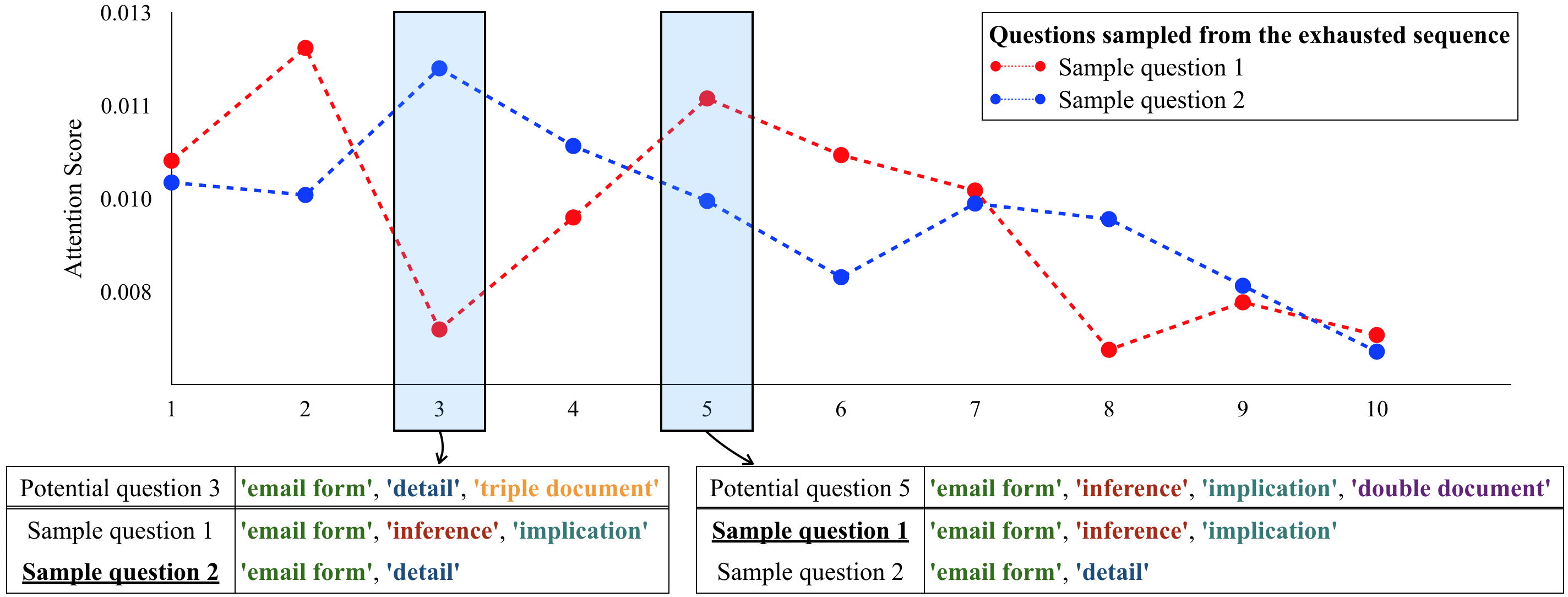}
\caption{The attention trajectory.}
\label{fig:attention_trajectory}
\end{figure*}

One attractive feature of our model is interpretability.
SantaTOEIC provides explanations for the reason why the questions are recommended. Figure \ref{fig:attention_trajectory} visualizes the attention trajectory. The attention score of two questions sampled from the exhausted sequence depends on the potential question. When potential question 3 in figure \ref{fig:attention_trajectory} is predicted, the attention score of sample question 2 with high tag matching ratio is higher than sample question 1. When potential question 5 is predicted, the order is opposite. Our results clearly show that higher tag matching ratios correspond to higher attention scores over the corresponding questions. Also, we quantitatively evaluate the performance of attention for all responses. We select the top-\textit{N} questions from the exhausted sequence based on attention scores and compute tag matching ratios between potential questions and the top \textit{k} questions. Table \ref{table:attention_top} shows that our learned attention model is consistent with domain experts.

\begin{table}[t]
\centering
\caption{The relationship attention and tag matching ratio}
\begin{tabular}{c||c}
\toprule 
Top-$\textit{N}$ & tag matching ratio (\%) \\
\hline
1&$\textbf{75.35}$\\
2&$\textbf{74.38}$\\
3&$\textbf{75.66}$\\
\hline
48&66.32\\
49&65.55\\
50&66.94\\
\bottomrule
\end{tabular}
\label{table:attention_top}
\end{table}

\subsection{Ablation Study}
We present an ablation study to illustrate the effect of Bi-LSTM and attention model choices.
First, Table \ref{table:variation_lstm} shows that the Bi-LSTM performs better than a vanilla LSTM and a fully-connected (FC) layer. For a fair comparison, the number of parameters of the FC-layer, LSTM, and Bi-LSTM are similarly set as: 2,142,210, 2,520,450, and 2,356,610. Second, Table \ref{table:variation_attention} shows dot-product and additive attentions show similar performance.

\begin{table}[t]
\centering
\caption{Ablation study on LSTM}
\begin{tabular}{c||ccc}
\toprule 
Metric  & FC layer  & LSTM  & Bi-LSTM   \\
\hline
F1-score& 0.8113  & 0.8109  & 0.8114\\
AUC & 0.7643  & 0.7647  & 0.7661\\
ACC & 0.7290  & 0.7292  & 0.7304\\
\bottomrule
\end{tabular}
\label{table:variation_lstm}
\end{table}

\begin{table}[t]
\centering
\caption{Ablation study on attention}
\begin{tabular}{c||ccc}
\toprule 
Metric  & w/o attention & dot-product   & additive    \\
\hline
F1-score& 0.8113  & 0.8113  &   0.8114  \\
AUC & 0.7659  & 0.7662  &   0.7661  \\
ACC & 0.7300  & 0.7305  &   0.7304  \\
\bottomrule
\end{tabular}
\label{table:variation_attention}
\end{table}

\section{Conclusion}
In this paper, we introduce a neural pedagogical agent for real-time user modeling of response correctness. We demonstrate improved and more efficient performance over existing methods and integrate our method into a smart review system which addresses characteristic problems of mobile education platform users. For future work, we plan to experiment with additional network architectures such as Transformers \cite{vaswani2017attention} as well as apply the principles from this paper to related downstream tasks. 

\bibliographystyle{ACM-Reference-Format}

\end{document}